\documentclass{article} 
\usepackage{arxiv, times}

\usepackage{amsmath,amsfonts,bm}

\def\eqref#1{equation~\ref{#1}}
\def\Eqref#1{Equation~\ref{#1}}

\def\1{\bm{1}}

\DeclareMathAlphabet{\mathsfit}{\encodingdefault}{\sfdefault}{m}{sl}
\SetMathAlphabet{\mathsfit}{bold}{\encodingdefault}{\sfdefault}{bx}{n}

\usepackage{hyperref}
\usepackage{url}
\usepackage{booktabs}
\usepackage{graphicx}

\title{Riccati State Space Models: \\Non-iterative Parallelization for Nonlinear Sequence Modeling}

\author{%
  Mónika Farsang\textsuperscript{\rm 1,2} \\
  \And
  Ramin Hasani\textsuperscript{\rm 2,3} \\
  \And
  Daniela Rus\textsuperscript{\rm 2} \\
  \And
  Radu Grosu\textsuperscript{\rm 1} \\
}

\begin{document}

\maketitle
\let\thefootnote\relax\footnotetext{\textsuperscript{\rm 1}TU Wien}
\let\thefootnote\relax\footnotetext{\textsuperscript{\rm 2}MIT CSAIL}
\let\thefootnote\relax\footnotetext{\textsuperscript{\rm 3}Liquid AI}
\let\thefootnote\relax\footnotetext{\textsuperscript{\rm }Corr. author: monika.farsang@tuwien.ac.at}

\begin{abstract}
State space models (SSMs) achieve efficient sequence processing because their affine state updates are closed under composition and can therefore be evaluated with an associative parallel scan. Nonlinear recurrent models can provide richer, state-dependent dynamics, but generally lose this compositional structure: parallel evaluation then requires iterative methods that repeatedly linearize and scan the recurrence. We ask, what state-dependent nonlinear dynamics can be designed to remain exactly composable? We answer by introducing RiccatiSSM, a nonlinear SSM, in which each state dimension follows an input-conditioned Riccati differential equation. Its quadratic state dependence makes the local Jacobian explicitly state-dependent, while its exact per-step flow under piecewise-constant inputs is a Möbius transformation. Since Möbius maps are closed under composition and compose through \(2\,{\times}\,2\) matrix multiplication, the complete nonlinear state trajectory can be evaluated exactly with a single associative parallel scan, without iterative linearization. We further derive a constrained parameterization that ensures bounded, contractive dynamics, and avoids poles in the fractional-linear state update. Across long-sequence classification, regression, and forecasting tasks, RiccatiSSM achieves competitive predictive performance while reducing runtime by $22{-}33\%$ compared to the nonlinear LrcSSM under matched architectures. These results demonstrate that state-dependent nonlinear dynamics can retain exact composability and be evaluated efficiently within a single parallel scan.
\end{abstract}

\section{Introduction}
State space models (SSMs) have emerged as efficient sequence models for long-context tasks, combining recurrent state updates with highly parallel training
\citep{gu2022efficientlymodelinglongsequences,smith2023simplified,gu2023mamba}.
A key source of this efficiency is the structure of their recurrence: although the transition parameters may depend on the input, the state update remains affine in the previous state. Affine maps are closed under composition, allowing an entire sequence to be evaluated using an associative parallel scan with linear work and logarithmic parallel depth.

This computational structure, however, restricts the dynamics that can be represented within the recurrent state update. In particular, the local state Jacobian of an affine recurrence does not depend on the current state. Nonlinear recurrent models can instead make the dynamics themselves state-dependent, allowing their local contraction rates and effective timescales to change along a trajectory. Liquid time-constant neural networks (LTC) provide a prominent example of this behavior by placing the current state inside the continuous-time neuron dynamics~\citep{hasani2021liquid}. Such state dependence increases the flexibility of the dynamics, but in general destroys the compositional structure of the model, that enables a single parallel scan.

Recent methods recover parallelism for nonlinear recurrences by treating sequence evaluation as a nonlinear trajectory-solving problem. DEER~\citep{limparallelizing}, for example, applies Newton iterations on the state trajectory, where each iteration linearizes the recurrence around the current estimate and solves the resulting affine recurrence in parallel. LrcSSM~\citep{farsang2025parallelization} applies this approach to biologically plausible liquid-resistance, liquid-capacitance recurrent dynamics, while designing the neuron dynamics such that they have a diagonal Jacobian, substantially reducing the cost of each iteration. Nevertheless, evaluating the nonlinear recurrence still requires repeated parallel scans, with the number of iterations determined by convergence.

In this work, we take a different approach: rather than starting from a general nonlinear recurrence and recovering parallelism through iterative linearization, we ask \textit{whether the nonlinear dynamics themselves can be chosen to remain exactly composable}. This requires a nonlinear family whose time-step maps are closed under composition with a fixed-size representation. We show that input-conditioned Riccati dynamics provide such a construction. Their exact per-step flow maps under piecewise-constant inputs are fractional-linear, or Möbius transformations, which compose through multiplication of \(2\times2\) matrices. Consequently, the complete nonlinear state trajectory can be evaluated within a single associative parallel scan, without Newton or fixed-point iterations.

\textit{This leads us to RiccatiSSM}, a diagonal nonlinear state space model, in which each state dimension follows an input-conditioned Riccati differential equation. The quadratic state term makes the local Jacobian explicitly state-dependent, preserving the central dynamical property motivating liquid recurrent models. Moreover, the Riccati structure makes the exact flow compositionally closed. Since unrestricted Riccati equations can exhibit finite-time divergence, we further derive a constrained parameterization that provides stable, bounded dynamics while preserving the exact Möbius flow. RiccatiSSM therefore occupies a middle ground between affine SSMs, which admit direct parallel scans but lack state-dependent local dynamics, and general nonlinear recurrent models, which provide such dynamics but require iterative methods for parallel evaluation.

\textbf{Our main contributions in this paper are as follows:}
\begin{itemize}
    \item\textit{We introduce RiccatiSSM}, a nonlinear state-dependent sequence
    model whose discrete-time flow maps are Möbius transformations. Their closure under composition enables exact, non-iterative sequence
    evaluation with a single associative scan.

    \item\textit{We derive a constrained Riccati parameterization} that makes these Riccati dynamics particularly suitable for sequence modeling by ensuring a bounded invariant state interval, contractive transitions, and a pole-free projective readout.

    \item\textit{We evaluate RiccatiSSM} on long-sequence classification, regression, and forecasting tasks, where it achieves competitive predictive performance while reducing runtime by \(22-33\%\) relative to iterative nonlinear LrcSSM under matched architectures.
\end{itemize}

\section{Background and Motivation}
\label{sec:background}

\subsection{Parallel scans and state-dependent dynamics}
\label{sec:scan_background}

Modern SSMs achieve efficient sequence evaluation because their discrete-time recurrence is affine in the previous state,
\begin{equation}
x_t = \Lambda_t x_{t-1} + b_t,
\label{eq:affine_recurrence}
\end{equation}
where $\Lambda_t$ and $b_t$ may depend on the current input $u_t$ but not on $x_{t-1}$. Each time step therefore defines an affine map $f_t(x)=\Lambda_t x+b_t$. Affine maps are closed under composition:
\begin{equation}
f_2(f_1(x))
=
\Lambda_2\Lambda_1x
+
\Lambda_2b_1+b_2,
\end{equation}
so a pair $(\Lambda_t,b_t)$ provides a fixed-size representation of each update and can be combined with an associative operator. The complete sequence can consequently be evaluated using a parallel prefix scan with $O(T)$ work and $O(\log T)$ parallel depth.

Allowing the dynamics to depend nonlinearly on the current state changes the recurrent update to:
\begin{equation}
x_t = f_t(x_{t-1}),
\label{eq:nonlinear_recurrence}
\end{equation}
where $f_t$ is nonlinear.
For a general nonlinear family, composing two updates does not yield another map with the same fixed-size representation. For example, composing two degree-$k$ polynomials generally produces a degree-$k^2$ polynomial. The representation therefore grows under composition, preventing the recurrence from being evaluated by a single fixed-size associative scan.

Methods such as DEER~\citep{limparallelizing} recover parallelism for general nonlinear recurrences through Newton iterations over the full state trajectory. At each iteration, the nonlinear recurrence is linearized, which is then evaluated by a parallel scan, but several such iterations may be required. Thus, general nonlinear state dependence can be parallelized, but at the cost of repeated scans. LrcSSM~\citep{farsang2025parallelization} applies this approach to liquid recurrent dynamics while designing the model to have a diagonal Jacobian, reducing the cost of each iteration.

\subsection{Closure under composition}
\label{sec:composition_background}

An alternative is to restrict the nonlinear update family itself. For a recurrence to admit exact evaluation by a single associative scan, we seek a family $\mathcal{F}$ whose elements have a fixed-size representation and are closed under composition:
\begin{equation}
f_1,f_2\in\mathcal{F}
\quad\Longrightarrow\quad
f_2\circ f_1\in\mathcal{F}.
\label{eq:composition_closure}
\end{equation}
Affine maps satisfy this property, but they remain linear in the state. We instead seek a nonlinear family that preserves the same closure property.

For a scalar state, fractional-linear (rational-linear), or Möbius, maps provide such a family:
\begin{equation}
f(x)=\frac{ax+b}{cx+d}.
\label{eq:mobius_background}
\end{equation}
Each such map can be represented projectively by a $2\times2$ matrix:
\begin{equation}
M_f =
\begin{pmatrix}
a & b\\
c & d
\end{pmatrix},
\end{equation}
and the composition of two Möbius maps corresponds to matrix multiplication:
\begin{equation}
M_{f_2\circ f_1} = M_{f_2}M_{f_1},
\label{eq:mobius_composition_background}
\end{equation}

Since matrix multiplication is associative, a sequence of Möbius updates can be composed exactly with a parallel prefix scan while retaining a fixed $2\times2$ representation at every step.

This observation motivates RiccatiSSM. Rather than approximating a general nonlinear recurrence through repeated linearization, we construct the state dynamics such that their exact discrete-time flow is a Möbius map. Riccati differential equations generate such fractional-linear flow maps, providing the continuous-time dynamics underlying RiccatiSSM, which we introduce next in Section~\ref{sec:riccatissm}.

\section{Riccati State Space Model}
\label{sec:riccatissm}

\begin{table}[t]
\centering
\caption{RiccatiSSM compared with DEER/ELK-style~\citep{limparallelizing,gonzalez2024towards} parallelization of general nonlinear
recurrences. Both evaluate state-dependent dynamics with associative scans,
but DEER/ELK approximate the general nonlinear update by Newton iteration updates until convergence, whereas RiccatiSSM restricts the dynamics to a family that composes
exactly in a single step. $T$ is the sequence length, $D$ the number of channels, and $K$ the
number of sweeps.}
\label{tab:vs_deer}
\small
\begin{tabular}{lll}
\toprule
 & DEER/ELK-based models & RiccatiSSM \\
\midrule
Dynamics                & any $f(x_{t-1},u_t)$        & Riccati: $\alpha+\beta x+\gamma x^{2}$ \\
Scan element            & affine ($(J_t, c_t)$, $J_t$ the state Jacobian)       & fractional-linear, $M_t \in \mathbb{R}^{2\times2}$ \\
Composition             & approximate (per iteration)  & exact \\
Sweeps per layer        & until convergence ($K$, data-dependent)         & $1$, fixed \\
Depth per layer         & $\mathcal{O}(K\log T)$       & $\mathcal{O}(\log T)$ \\
Work per layer          & $\mathcal{O}(K\,TD)$         & $\mathcal{O}(TD)$ \\
Within-step integration & based on the model              & exact (matrix exponential) \\
Convergence             & data-dependent, may stall    & not applicable \\
\bottomrule
\end{tabular}
\end{table}

We now construct a nonlinear sequence layer whose state-dependent
dynamics admit exact parallel evaluation. The construction has four
steps: (1) We first define an input-conditioned Riccati vector field.
(2) We lift this nonlinear scalar ODE to a two-dimensional linear
system, whose exact zero-order-hold solution induces a Möbius state
update. (3) Since Möbius maps compose through matrix multiplication, the full
sequence can be evaluated with a single associative scan. (4) Finally, we
constrain the coefficients to ensure bounded trajectories and
contractive state transitions.

\begin{figure}
    \centering
    \includegraphics[width=0.8\linewidth]{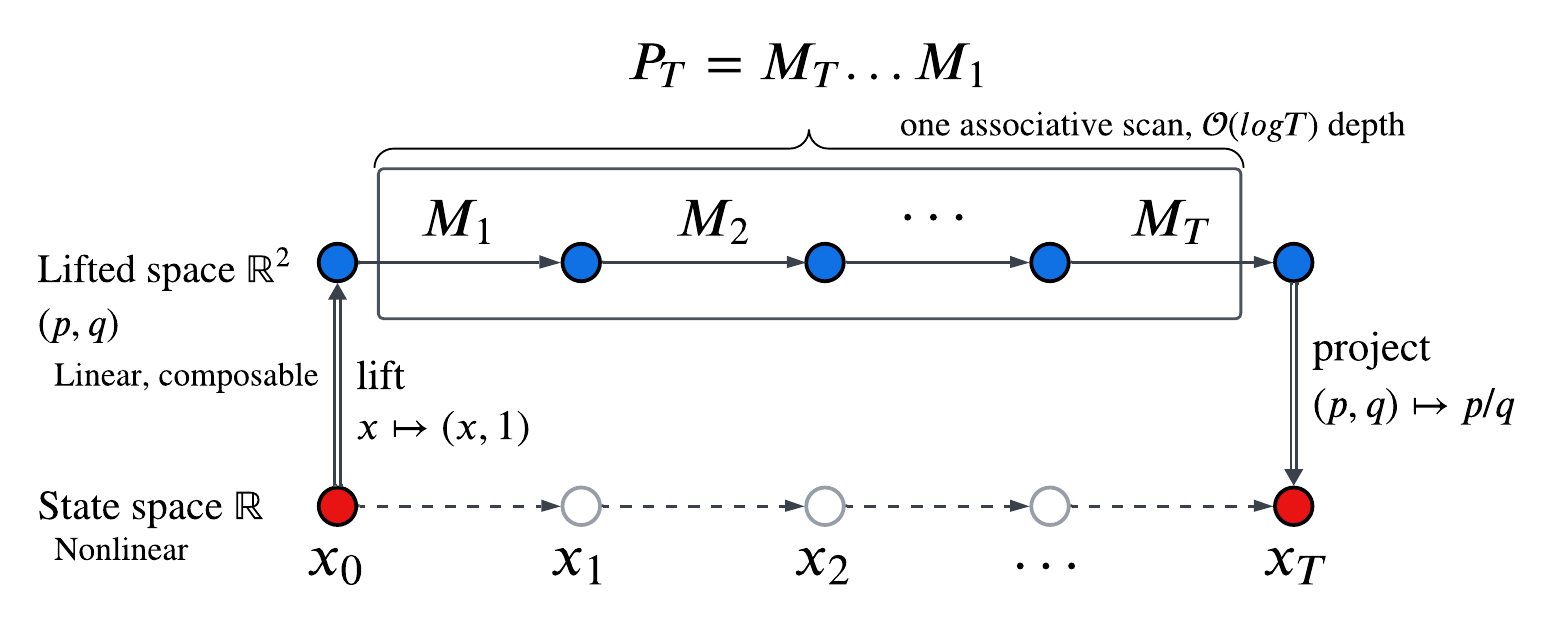}
    \caption{RiccatiSSM evaluates a nonlinear recurrence by a detour through a two-dimensional linear one. Bottom (grey, dashed): the states $x_t$ of one channel, whose Möbius updates are nonlinear in the state. Composing such maps directly requires clearing nested fractions, so the bottom row offers no cheap merge to scan over. Top (blue, solid): the
  same states as homogeneous pairs $(p_t,q_t)$, where each step is the matrix
  $M_t$. The initial state is lifted once, $x_0 \mapsto (x_0,1)$, and states
  are recovered by $x_t = p_t/q_t$. Matrix multiplication is associative, so
  the top row collapses into prefix products $P_t = M_t \cdots M_1$ under one
  associative scan of depth $\mathcal{O}(\log T)$, with no fixed-point
  iterations; all $D$ channels scan in parallel in the same round.}
    \label{fig:main_figure}
\end{figure}

\subsection{Input-dependent Riccati dynamics}
\label{sec:model}

Let \(u_t\) denote the input at time \(t\), and let
\(x_t \in \mathbb{R}^{D}\) denote the hidden state. As in diagonal
state space models and LrcSSM \citep{farsang2025parallelization}, RiccatiSSM
uses independent scalar dynamics across state dimensions; interactions
between dimensions are handled by the surrounding input and output
projections. For each dimension \(i\), we define the continuous-time
dynamics as:
\begin{equation}
    \dot{x}_i
    =
    \varepsilon_i(u)
    \left[
        \alpha_i(u)
        +
        \beta_i(u)x_i
        +
        \gamma_i(u)x_i^2
    \right].
    \label{eq:riccati_ode}
\end{equation}
The coefficients
\(\alpha_i(u)\), \(\beta_i(u)\), \(\gamma_i(u)\), and
\(\varepsilon_i(u)\) are produced by input-dependent parameter heads.
Here, \(\alpha_i\) controls the input drive, \(\beta_i\) the linear state dependence,
\(\gamma_i\) the quadratic curvature. Inspired by LRCs~\citep{farsangliquid}, we use
\(\varepsilon_i \in (0,1)\) to control the overall speed of the dynamics.

Unlike in an input-varying affine SSM, the local state Jacobian of the dynamics described in
\Eqref{eq:riccati_ode}, depends explicitly on the current state:
\begin{equation}
    \frac{\partial \dot{x}_i}{\partial x_i}
    =
    \varepsilon_i
    \left(
        \beta_i + 2\gamma_i x_i
    \right).
    \label{eq:liquid_tc}
\end{equation}
Thus, the local contraction rate of the dynamics can vary with both the input, through the coefficients $\varepsilon$, $\beta$ and $\gamma$, and the current state, through the quadratic term, giving the dynamics a liquid, input-and-state dependent character, like in~\cite{hasani2021liquid,farsangliquid}. 

When \(\gamma_i=0\), \Eqref{eq:riccati_ode} reduces to an input-varying
linear ODE. The quadratic term is therefore the lowest-degree
polynomial term that introduces explicit state-dependence into the local Jacobian.

\subsection{Projective lift and exact discretization}
\label{sec:discretization}

A scalar Riccati equation can be represented as the ratio of two
coordinates evolving under a linear system. For clarity, we suppress
the state index \(i\) and write
\[
    x = \frac{p}{q}.
\]
Define the lifted state \(z=(p,q)^\top\), and consider the linear
system:
\begin{equation}
    \frac{d}{dt}
    \begin{pmatrix}
        p \\ q
    \end{pmatrix}
    =
    \begin{pmatrix}
        \dot{p} \\ \dot{q}
    \end{pmatrix}    
    =
    L
    \begin{pmatrix}
        p \\ q
    \end{pmatrix},
    \qquad
    L
    =
    \varepsilon
    \begin{pmatrix}
        \beta/2 & \alpha \\
        -\gamma & -\beta/2
    \end{pmatrix}.
    \label{eq:lift}
\end{equation}
Applying the quotient rule to \(x=p/q\) gives:
\begin{equation}
    \dot{x}
    =
    \frac{\dot{p}q-p\dot{q}}{q^2}
    =
    \varepsilon
    \left(
        \alpha+\beta x+\gamma x^2
    \right),
    \label{eq:lift_verification}
\end{equation}
recovering \Eqref{eq:riccati_ode}. A full derivation of \Eqref{eq:riccati_ode} is given in
Appendix~\ref{app:lift}.

We apply a zero-order hold over each interval
\([t\Delta t,(t+1)\Delta t)\): the coefficients are computed from
\(u_t\) and held constant over the interval. The lifted system then has the exact
one-step solution:
\begin{equation}
    \begin{pmatrix}
        p_t \\ q_t
    \end{pmatrix}
    =
    M_t
    \begin{pmatrix}
        p_{t-1} \\ q_{t-1}
    \end{pmatrix},
    \qquad
    M_t = \exp(\Delta t\,L_t).
    \label{eq:step_matrix}
\end{equation}
Since \(L_t\) is traceless and satisfies
\(L_t^2=\omega_t^2 I\), its exponential has the closed form:
\begin{equation}
    M_t
    =
    \cosh(\omega_t\Delta t)\,I
    +
    \frac{\sinh(\omega_t\Delta t)}{\omega_t}\,L_t,
    \qquad
    \omega_t
    =
    \sqrt{
        \varepsilon_t^2
        \left(
            \frac{\beta_t^2}{4}-\alpha_t\gamma_t
        \right)
    }.
    \label{eq:expm}
\end{equation}

Our stability parameterization that we introduce below, makes sure that
\(\omega_t>0\), so the hyperbolic form in \Eqref{eq:expm} is always
well defined. Writing:
\[
    M_t =
    \begin{pmatrix}
        a_t & b_t \\
        c_t & d_t
    \end{pmatrix},
\]
the induced update of the original scalar state becomes:
\begin{equation}
    x_t
    =
    \frac{a_t x_{t-1}+b_t}{c_t x_{t-1}+d_t}.
    \label{eq:riccati_moebius_step}
\end{equation}
Thus, exact zero-order-hold integration of a Riccati ODE produces a
Möbius update. This is the property that makes the nonlinear recurrence
scan-compatible. Although the dynamics are linear in the lifted
coordinates, the projected state ($x = p/q$) follows the nonlinear,
state-dependent Riccati dynamics in \Eqref{eq:riccati_ode}. We discuss
this and its relation to liquid dynamics further in
Appendix~\ref{remark:linearlift_nonlineardynamics}.

\subsection{Parallel sequence evaluation}
\label{sec:scan}

The Möbius updates in \Eqref{eq:riccati_moebius_step} compose directly through their matrix representations. Define the the following prefix product:
\begin{equation}
    P_t
    =
    M_t M_{t-1}\cdots M_1
    =
    \begin{pmatrix}
        P_t^{(11)} & P_t^{(12)} \\
        P_t^{(21)} & P_t^{(22)}
    \end{pmatrix}.
    \label{eq:prefix_product}
\end{equation}
Since matrix multiplication is associative, all prefix products
\(\{P_t\}_{t=1}^{T}\) can be computed with a single parallel scan. The
state at every time step is then recovered from the initial state
\(x_0\) as:
\begin{equation}
    x_t
    =
    \frac{
        P_t^{(11)}x_0+P_t^{(12)}
    }{
        P_t^{(21)}x_0+P_t^{(22)}
    }.
    \label{eq:readout}
\end{equation}

The forward pass therefore consists of three non-iterative stages:
\begin{enumerate}
    \item \textit{Coefficient generation.} Compute the coefficients
    \((\alpha_t,\beta_t,\gamma_t,\varepsilon_t)\) and the corresponding matrices \(M_t\), for all
    time steps and state dimensions, in parallel.
    \item \textit{Parallel scan.} Compute the matrix prefix-products in
    \Eqref{eq:prefix_product}, by using the associativity property of \(2\times2\) matrix
    multiplication.
    \item \textit{Projective readout.} Recover the state \(x_t\) from the readout of
    \Eqref{eq:readout}.
\end{enumerate}
It is important to note that for a diagonal state dynamics of dimension \(D\) and sequence length \(T\),
the recurrent computation requires \(\mathcal{O}(TD)\) work and
\(\mathcal{O}(\log T)\) parallel depth. Under the zero-order-hold
assumption, each Riccati step is integrated exactly. Hence the scan
introduces neither iterative approximation nor a within-step numerical
integration error.

\subsection{Stable parameterization}
\label{sec:stability_main}

Unfortunately, a general Riccati equation can diverge in finite time. In the
fractional-linear update of \Eqref{eq:riccati_moebius_step}, this 
divergence corresponds to the denominator approaching zero (a pole). We avoid
this behavior by constraining the Riccati coefficients in such a way that the
dynamics admit a bounded invariant interval, and the state transition is
contractive within this interval. The resulting parameterization and
stability conditions are detailed in Appendix~\ref{sec:stability}.

\subsection{Relation to Liquid Dynamics}
\label{sec:correspondence}

RiccatiSSM and LrcSSM both use diagonal continuous-time neuron
dynamics, in order to avoid the matrix-multiplication blow-up 
within the associative scan. However, they differ in how they represent state dependence. LrcSSM
places the state inside saturating nonlinearities, making its update, leak, and speed (elastance) factor state-dependent. RiccatiSSM
instead uses input-dependent coefficients and represents state
dependence explicitly through a quadratic term. This structure yields an exact
fractional-linear step map that can be composed with a single
associative scan.

We establish a local correspondence between the two models by
constructing a second-order Taylor approximation of the LRC vector
field with respect to the state around its zero value. We further evaluate the accuracy of
this local approximation and distinguish it from the discretization
error introduced by the explicit-Euler update used in LrcSSM. The
derivation and the matched-dynamics experiments are provided in Appendix~\ref{sec:taylor_analysis} and~\ref{sec:dyn_setup}.

\section{Experiments}
\label{sec:experiments}


We evaluate RiccatiSSM on long-sequence classification, regression, and forecasting tasks, comparing its predictive performance with transformer-based architectures, neural differential equation models, and with linear, oscillatory, input-selective and nonlinear SSMs. We then measure the computational benefit of replacing the iterative quasi-DEER evaluation of LrcSSM with the exact single-scan evaluation of RiccatiSSM. Finally, we ablate the RiccatiSSM parameterization to study the effects of input-dependent coefficients and the connection to LRC dynamics.

\subsection{Experimental setup}
\label{sec:exp_setup}

RiccatiSSM follows the sensory-synapse input encoding of LrcSSM~\citep{farsang2025parallelization}, but uses independent learned projections of this representation to produce the Riccati coefficients $\alpha(u)$, $\beta(u)$, $\gamma(u)$, and $\varepsilon(u)$. The resulting coefficients are mapped through the constrained parameterization of Section~\ref{sec:stability}, which enforces the stability conditions of RiccatiSSM. We separately evaluate an LRC-tied parameterization in Section~\ref{sec:ablations}, where these coefficients are constrained by the Taylor-matching relations derived in Appendix~\ref{sec:taylor_analysis}.

\begin{table}[h]
    \centering
        \caption{Test accuracy comparison of different models. The performance of the models marked by \dag{} is reported from~\cite{rusch2024oscillatory}, those with \textdaggerdbl~from~\cite{moreno2024}, and $\diamond$~from~\cite{farsang2025parallelization}. The same hyperparameter tuning protocol and dataset splitting over the same 5 seeds were used. Bold highlights the top 3 models.
        }
    \tiny
    \resizebox{\textwidth}{!}{%
    \begin{tabular}{lcccccc}
        \toprule
        & Heart & SCP1 & SCP2  &  Ethanol & Motor & Worms  \\
        Sequence length & 405 & 896 & 1,152& 1,751 & 3,000 & 17,984  \\
        Input size & 61 & 6 &  7& 2 & 63 & 6\\
        \#Classes & 2 & 2 & 2 & 4 & 2 & 5\\
        \midrule
        Transformer\textsuperscript{\textdaggerdbl} & 70.5 $\pm$ 0.1 & 84.3 $\pm$ 6.3 & 49.1 $\pm$ 2.5 & \textbf{40.5 $\pm$ 6.3} & 50.5 $\pm$ 3.0 & OOM \\
        RFormer\textsuperscript{\textdaggerdbl} & 72.5 $\pm$ 0.1  & 81.2 $\pm$ 2.8 & 52.3 $\pm$ 3.7 & 34.7 $\pm$ 4.1  & 55.8 $\pm$ 6.6 & \textbf{90.3 $\pm$ 0.1} \\
        \midrule
        NRDE\textsuperscript{\dag} & 73.9 $\pm$ 2.6 &  76.7 $\pm$ 5.6 & 48.1 $\pm$ 11.4 & 31.4 $\pm$ 4.5 & 54.0 $\pm$ 7.8 & 77.2 $\pm$ 7.1  \\
        NCDE\textsuperscript{\dag} & 68.1 $\pm$ 5.8 &  80.0 $\pm$ 2.0 & 49.1 $\pm$ 6.2  & 22.0 $\pm$ 1.0 & 51.6 $\pm$ 6.2 & 62.2 $\pm$ 2.2 \\
        Log-NCDE\textsuperscript{\dag} & 74.2 $\pm$ 2.0 & 82.1 $\pm$ 1.4 & 54.0 $\pm$ 2.6 & 35.9 $\pm$ 6.1 & 57.2 $\pm$ 5.6 & 82.8 $\pm$ 2.7  \\
        \midrule
        LRU\textsuperscript{\dag} & \textbf{78.1 $\pm$ 7.6} & \ 84.5 $\pm$ 4.6 & 47.4 $\pm$ 4.0  & 23.8 $\pm$ 2.8 & 51.9 $\pm$ 8.6 & 85.0 $\pm$ 6.2 \\
        S5\textsuperscript{\dag} & 73.9 $\pm$ 3.1 &  87.1 $\pm$ 2.1 & \textbf{55.1 $\pm$ 3.3} & 25.6 $\pm$ 3.5 & 53.0 $\pm$ 3.9 & 83.9 $\pm$ 4.1   \\
        LinOSS-IMEX\textsuperscript{\dag} & 75.5 $\pm$ 4.3 & \textbf{87.5 $\pm$ 4.0} & \textbf{58.9 $\pm$ 8.1}  & 29.9 $\pm$ 1.0 & 57.9 $\pm$ 5.3 & 80.0 $\pm$ 2.7  \\
        LinOSS-IM\textsuperscript{\dag} &  75.8 $\pm$ 3.7 & \textbf{87.8 $\pm$ 2.6} & \textbf{58.2 $\pm$ 6.9} & 29.9 $\pm$ 0.6 & \textbf{60.0 $\pm$ 7.5} & \textbf{95.0 $\pm$ 4.4}    \\
        Mamba\textsuperscript{\dag} & \textbf{76.2 $\pm$ 3.8} & 80.7 $\pm$ 1.4 & 48.2 $\pm$ 3.9 & 27.9 $\pm$ 4.5 & 47.7 $\pm$ 4.5  & 70.9 $\pm$ 15.8\\
        S6\textsuperscript{\dag} &  \textbf{76.5 $\pm$ 8.3} & 82.8 $\pm$ 2.7 & 49.9 $\pm$ 9.4 & 26.4 $\pm$ 6.4 & 51.3 $\pm$ 4.7 & 85.0 $\pm$ 16.1  \\
        \midrule
        LrcSSM\textsuperscript{$\diamond$}  & 72.7 $\pm$ 5.7 & 85.2 $\pm$ 2.1 & 53.9 $\pm$ 7.2 & \textbf{36.9 $\pm$ 5.3} & \textbf{58.6 $\pm$ 3.1} & \textbf{90.6 $\pm$ 1.4}\\
        RiccatiSSM (ours) & 73.0 $\pm$ 6.8 & \textbf{87.4 $\pm$ 4.3} & 53.2 $\pm$ 4.3& \textbf{36.6 $\pm$ 3.4}& \textbf{59.6 $\pm$ 3.1}& 83.9 $\pm$ 6.4 \\
        \bottomrule
    \end{tabular}%
    }
    \label{tab:performance_comparison}
\end{table}

Results are reported over five random seeds. Datasets, model configurations, and hyperparameter ranges are provided in Appendix~\ref{app:experiments}, with the complete hyperparameter settings.

\subsection{Long-sequence benchmark results}
\label{sec:benchmark_results}
\paragraph{Classification.} We first evaluate RiccatiSSM on six multivariate time-series classification datasets from the UEA archive, spanning sequence lengths from $405$ to $17{,}984$. Table~\ref{tab:performance_comparison} compares RiccatiSSM with continuous-time recurrent models, linear SSMs, Transformer-based architectures, and the nonlinear LrcSSM baseline. RiccatiSSM is competitive across datasets and closely matches LrcSSM while avoiding its iterative sequence evaluation. In particular, it improves over LrcSSM on SCP1 and MotorImagery, while obtaining similar performance on Heartbeat, SCP2, and EthanolConcentration. On EigenWorms, RiccatiSSM shows higher variability across seeds, with individual runs ranging from 77.8\% to 91.7\%, suggesting greater sensitivity to optimization on this dataset.

\paragraph{Long-range Regression and Forecasting}

We next evaluate regression on PPG-DaLiA~\citep{reiss2019deep}. Following the same tuning protocol, RiccatiSSM achieves an MSE of $7.15\pm1.01\times10^{-2}$ (Table~\ref{tab:ppg}), improving over LrcSSM ($10.89\pm0.96\times10^{-2}$) and remaining competitive with the strongest linear SSM baseline, the LinOSS.

\begin{table}[h]
\centering
\caption{Mean squared error (MSE × $10^{-2}$) for different models on the PPG-DaLiA dataset. The performance of the models marked by \dag{} is reported from~\cite{rusch2024oscillatory} and $\diamond$~from~\cite{farsang2025parallelization}. All results are averaged over 5 seeds.}
\small
\begin{tabular}{lc}
\toprule
\textbf{Model} & MSE $\times 10^{-2}$ ($\downarrow$)\\
\midrule
NRDE\textsuperscript{\dag}   \citep{morrill2021neural}        & 9.90 $\pm$ 0.97 \\
NCDE\textsuperscript{\dag}   \citep{kidger2020neural}        & 13.54 $\pm$ 0.69 \\
Log-NCDE\textsuperscript{\dag}   \citep{walkerlog}    & 9.56 $\pm$ 0.59 \\
\midrule
LRU\textsuperscript{\dag}    \citep{orvieto2023resurrecting}        & 12.17 $\pm$ 0.49 \\
S5\textsuperscript{\dag}     \citep{smith2023simplified}        & 12.63 $\pm$ 1.25 \\
LinOSS-IMEX\textsuperscript{\dag} \citep{rusch2024oscillatory}   & \textbf{7.50 $\pm$ 0.46} \\
LinOSS-IM\textsuperscript{\dag}   \citep{rusch2024oscillatory}   & \textbf{6.40 $\pm$ 0.23} \\
S6\textsuperscript{\dag}  \citep{gu2023mamba}          & 12.88 $\pm$ 2.05 \\
Mamba\textsuperscript{\dag}  \citep{gu2023mamba}           & 10.65 $\pm$ 2.20 \\
\midrule
LrcSSM\textsuperscript{$\diamond$}    \citep{farsang2025parallelization}     & 10.89 $\pm$ 0.96 \\
RiccatiSSM (ours) & \textbf{7.15 $\pm$ 1.13} \\
\bottomrule
\end{tabular}
\label{tab:ppg}
\end{table}

To evaluate long-horizon forecasting, we use the Weather dataset with an input context of $720$ time steps and predict the subsequent $720$ steps, following~\citet{zhou2021informer}. As shown in Table~\ref{tab:weather}, RiccatiSSM obtains a mean absolute error of $0.5681$, improving over LrcSSM ($0.5888$) and several recurrent, Transformer-based, and state-space baselines, while remaining close to LinOSS.

\begin{table}[h]
\centering
\caption{Mean absolute error on the weather dataset predicting 720 future time steps based on 720 past time steps. The performance of the models marked by \dag{} is reported from~\cite{rusch2024oscillatory}.}
\small
\begin{tabular}{lc}
\toprule
\textbf{Model} & Mean Absolute Error ($\downarrow$) \\
\midrule
Informer\textsuperscript{\dag} \citep{zhou2021informer}& $0.731$ \\
LogTrans\textsuperscript{\dag} \citep{li2019enhancing}& $0.773$ \\
Reformer\textsuperscript{\dag} \citep{kitaev2020reformer}& $1.575$ \\
\midrule
LSTMa\textsuperscript{\dag} \citep{bahdanau2014neural}& $1.109$ \\
LSTnet\textsuperscript{\dag} \citep{lai2018modeling} & $0.757$ \\
\midrule
S4\textsuperscript{\dag} \citep{gu2022efficientlymodelinglongsequences}&$ 0.5783$ \\
LinOSS-IMEX\textsuperscript{\dag}  \citep{rusch2024oscillatory}& $\mathbf{0.5081}$ \\
LinOSS-IM\textsuperscript{\dag}  \citep{rusch2024oscillatory}&  $\mathbf{0.5282}$ \\
\midrule
LrcSSM \citep{farsang2025parallelization} & 0.5888 \\
RiccatiSSM (ours) & $\mathbf{0.5681}$  \\
\bottomrule
\end{tabular}
\label{tab:weather}
\end{table}

\subsection{Runtime and parallel-evaluation cost}
\label{sec:runtime}

We next evaluate whether eliminating iterative nonlinear sequence evaluation translates into an end-to-end computational benefit. We compare RiccatiSSM with LrcSSM using the same 6-layer architecture and a state size of $64$. While RiccatiSSM evaluates its recurrent dynamics with a single associative scan, LrcSSM requires approximately three quasi-DEER iterations, each involving a scan of a linearized recurrence.

\begin{figure}
\centering
\includegraphics[width=0.5\linewidth]{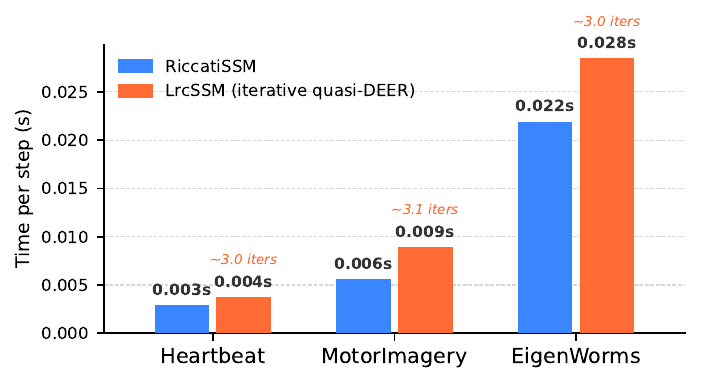}
\caption{Time per step (seconds) for RiccatiSSM and LrcSSM across three UEA time-series classification datasets using the same architecture on an NVIDIA A100 GPU. RiccatiSSM evaluates each layer with a single parallel scan, whereas LrcSSM requires approximately three quasi-DEER iterations. The runtime gap is largest for the longer EigenWorms sequences.}
\label{fig:timing_comparison}
\end{figure}

Figure~\ref{fig:timing_comparison} shows that eliminating these iterative solves consistently reduces wall-clock runtime. The reduction is smaller than the approximately $3\times$ difference in the number of scans because sequence evaluation constitutes only part of the complete architecture: embeddings, MLPs, normalization layers, and other operations are evaluated once in both models. Across the three datasets, RiccatiSSM requires $67-78\%$ of the LrcSSM runtime, corresponding to a $22-33\%$ reduction in end-to-end computation. The difference is largest on EigenWorms, which has the longest sequences. Additional runtime measurements are provided in Table~\ref{tab:runtime} in the Appendix.

\subsection{Ablations}
\label{sec:ablations}
\paragraph{Input-dependent coefficients.}
We ablate several components of the RiccatiSSM parameterization using PPG-DaLiA. First, we remove the input-dependent speed factor $\varepsilon(u)$, which increases the test MSE by $1.3\times10^{-2}$. We then restrict $\beta$, $\gamma$, and $\varepsilon$ to be input-independent, leaving only $\alpha(u)$ to inject the input into the state dynamics. Note that at least one of the coefficients should remain input dependent, such that the input can influence the state dynamics. The results indicate that input-dependent modulation of the full Riccati dynamics is beneficial.

\paragraph{Free vs. LRC-tied parameterization.}
The local correspondence derived in Appendix~\ref{sec:taylor_analysis} provides a second way to parameterize RiccatiSSM. Rather than learning the Riccati coefficients independently, we constrain them according to the Taylor-matching relations derived from the LrcSSM vector field. This LRC-tied parameterization obtains an MSE of $8.47\pm0.47\times10^{-2}$, compared with $7.15\pm1.01\times10^{-2}$ for the freely parameterized RiccatiSSM. Thus, retaining the LRC-derived coupling is not necessary for RiccatiSSM and is more restrictive than learning the Riccati coefficients independently subject to the stability constraints.The reachable coefficient spaces induced by the free and LRC-tied parameterizations are visualized in Figure~\ref{fig:reachable_param} in the Appendix.

\begin{table}[t]
    \centering
    \small
    \setlength{\tabcolsep}{5pt}
    \renewcommand{\arraystretch}{1.15}
    \caption{Ablation of the RiccatiSSM parameterization on PPG-DaLiA.
    Test MSE ($\times 10^{-2}$, lower is better), reported as mean $\pm$ standard deviation over five seeds.}
    \label{tab:riccati_ablation}
    \begin{tabular}{lcc}
        \toprule
        Model & Input-dependent coefficients & MSE ($\downarrow$) \\
        \midrule
        \textbf{RiccatiSSM (ours)}
            & $\alpha(u),\beta(u),\gamma(u),\varepsilon(u)$
            & $\mathbf{7.15 \pm 1.01}$ \\
        $\varepsilon$ ablation
            & $\alpha(u),\beta(u),\gamma(u)$
            & $8.45 \pm 0.38$ \\
        Reduced input dependence
            & $\alpha(u),\beta,\gamma,\varepsilon$
            & $8.71 \pm 1.10$ \\
        RiccatiSSM (LRC-tied)
            & LRC Taylor-matched
            & $8.47 \pm 0.47$ \\
        \bottomrule
    \end{tabular}
\end{table}

\subsection{Limitations}
\label{sec:limitations}
Compared to general nonlinear systems, RiccatiSSM is restricted to a specific second-order polynomial form in which the coefficients can be input-dependent but not state-dependent, i.e., $\alpha$, $\beta$, and $\gamma$ depend only on the input. State-dependent nonlinearity instead enters explicitly through the quadratic curvature term $\gamma(u)x^2$. Additional constraints are required to ensure stable dynamics. Despite these restrictions, RiccatiSSMs define an interesting new family of nonlinear SSMs that can be parallelized exactly with a single scan without iterative methods.

\section{Related Work}
\label{sec:related}
\vspace{-1ex}
\paragraph{Parallel evaluation of nonlinear recurrences.}
Linear and input-dependent linear SSMs evaluate long sequences with
associative scans because affine state updates are closed under
composition \citep{smith2023simplified, gu2023mamba}. Several recent methods
extend parallel evaluation to nonlinear recurrences through iterative
linearization. DEER \citep{limparallelizing} formulates sequence evaluation
as a fixed-point problem and applies Newton-style iterations, where
each iteration scans a linearized recurrence. Quasi-DEER and ELK
\citep{gonzalez2024towards} develop related iterative schemes intended to
improve stability and scalability. Fixed-Point RNNs
\citep{movahedi2025fixedpoint} express dense linear recurrences as
fixed points of parallelizable diagonal systems, while
\citet{UnifyingFramework2026} study Newton, Picard, and Jacobi
approaches within a common successive-linearization framework.
\citet{gonzalez2025predictability} further relate the convergence of
these methods to properties such as contractivity of the underlying
dynamics. Here, we explore the boundary of associative scans.

\paragraph{Liquid and continuous-time models.}
LTC \citep{hasani2021liquid} makes each neuron's
effective time constant depend on it own state, by using a liquid conductance. CfC
\citep{hasani2022closed} removes the ODE solver by integrating a single
step in closed form, but the sequence recursion remains sequential and
its coefficients remain state-dependent, so parallelization over $T$
still requires DEER-style iteration. LRC and LrcSSM
\citep{farsangliquid,farsang2025parallelization} extend this family with a liquid capacitance, and scale it to long sequences with a
diagonal Jacobian and a contraction-based gradient-stability
guarantee. While not fully following the biological constraints anymore,
RiccatiSSM keeps both the state-dependent time constant
(\Eqref{eq:liquid_tc}) and the contraction guarantee, and additionally
makes the \emph{within-step} discretization exact rather than explicit
Euler.

\paragraph{Möbius maps and parallel prefix.}
That fractional-linear maps compose thruough a $2\times 2$ matrix
multiplication, and that first-order rational recurrences therefore
admit parallel-prefix evaluation, is classical
\citep{kogge1973parallel, kogge1974parallel}; we claim no novelty for the
primitive itself. In machine learning, Möbius transformations have
been used as feature maps in attention
\citep{halacheva2024expanding}, but as static transformations of
representations, not as temporal dynamics composed over time. The
closest use of the primitive as a sequence recurrence is Kalman Linear
Attention \citep{shaj2026kalman}, where the information-form
\emph{precision} of a linear-Gaussian latent follows a Möbius
recursion composed by an associative scan. The hidden \emph{state} evolves linearly and the fractional-linear map
acts on an uncertainty statistic. In RiccatiSSM the map acts on the
state itself (\Eqref{eq:riccati_ode}), which is what yields a
state-dependent time constant and places the model in the liquid
family.

\section{Conclusion}
\vspace{-1ex}
We introduced RiccatiSSM, a nonlinear sequence layer whose input-conditioned Riccati dynamics yield exact fractional-linear state updates. Because these updates are closed under composition through $2\times2$ matrix multiplication, RiccatiSSM evaluates the full sequence with a single associative parallel scan, avoiding the iterative Newton updates required by general nonlinear recurrences. A constrained parameterization keeps the dynamics bounded and contractive while retaining state-dependent timescales. Across long-sequence classification, regression, and forecasting tasks, RiccatiSSM achieves competitive predictive performance, while reducing end-to-end runtime by $22-33\%$ relative to the iterative LrcSSM implementation. More broadly, RiccatiSSM shows that state-dependent nonlinear dynamics and exact parallel evaluation need not be mutually exclusive: restricting the dynamics to a nonlinear family closed under composition provides a middle ground between linear SSMs and general nonlinear recurrent models.

\subsubsection*{Acknowledgments}
Research was sponsored by the Department of the Air Force Artificial Intelligence Accelerator and was accomplished under Cooperative Agreement Number FA8750-19-2-1000. The views and conclusions contained in this document are those of the authors and should not be interpreted as representing the official policies, either expressed or implied, of the Department of the Air Force or the U.S. Government. The U.S. Government is authorized to reproduce and distribute reprints for Government purposes notwithstanding any copyright notation herein.

\bibliographystyle{plainnat}
\bibliography{references}

\newpage
\appendix

\section{Additional Theoretical Results}
\label{app:theory}

This appendix proves the algebraic and stability properties underlying
RiccatiSSM. We first show the parametrization, then that a Riccati equation is the projective image of a two-dimensional linear system. We then derive the exact 
matrix exponential used for zero-order-hold discretization, establish
closure of the resulting Möbius updates under composition, and prove
the guarantees of the constrained parameterization.

\subsection{Stable parameterization}
\label{sec:stability}

A general Riccati equation can diverge in finite time. In the
fractional-linear update of \Eqref{eq:riccati_moebius_step}, such a 
divergence corresponds to the denominator approaching zero. We avoid this behavior by
parameterizing the coefficients such that a fixed interval \([-B,B]\), with \(B>0\), is forward-invariant and the dynamics are contractive within this interval. 

For readability, we suppress the state-dimension and time indices in
this subsection. At each time step, the coefficient network produces
four unconstrained real-valued outputs
\begin{equation}
    (\hat{\alpha},\hat{\beta},\hat{\gamma},\hat{\varepsilon})
    \in \mathbb{R}^{4}.
    \label{eq:raw_coefficient_outputs}
\end{equation}
The hats indicate raw network outputs before applying the constraints.
We first define
\begin{equation}
    \varepsilon
    =
    \sigma(\hat{\varepsilon}),
    \qquad
    s
    =
    s_{\min}+\operatorname{softplus}(\hat{\beta}),
    \qquad
    \gamma
    =
    \hat{\gamma},
    \label{eq:stable_intermediate_parameters}
\end{equation}
where \(s_{\min}>0\) is a fixed minimum contraction margin. The
remaining Riccati coefficients are then parameterized as
\begin{equation}
    \alpha
    =
    \left(sB+|\gamma|B^2\right)\tanh(\hat{\alpha}),
    \qquad
    \beta
    =
    -\left(s+2|\gamma|B\right).
    \label{eq:stable_parameterization}
\end{equation}
Thus, \(\hat{\alpha}\) determines the bounded drive coefficient \(\alpha\), \(\hat{\beta}\) determines the positive contraction margin
\(s\), \(\hat{\gamma}\) determines the curvature \(\gamma\), and
\(\hat{\varepsilon}\) determines the speed factor
\(\varepsilon\in(0,1)\).

In particular, \(\hat{\beta}\) does not directly parameterize the
linear Riccati coefficient \(\beta\). Instead, it determines the positive margin
\(s\), from which \(\beta\) is constructed to offset
the largest possible state-dependent contribution \(2|\gamma|x\) to the Jacobian over
\(x\in[-B,B]\), while retaining a contraction margin of at least $s$. The resulting coefficients satisfy,
\begin{equation}
    |\alpha|
    \leq
    sB+|\gamma|B^2,
    \qquad
    \beta
    \leq
    -s
    \leq
    -s_{\min}.
    \label{eq:coefficient_bounds}
\end{equation}

These constraints give three properties.

\paragraph{Bounded trajectories.}
Consider the scalar Riccati vector field without its positive speed
factor,
\begin{equation}
    g(x)
    =
    \alpha+\beta x+\gamma x^2.
    \label{eq:riccati_vector_field}
\end{equation}
At the upper boundary \(x=B\), using~\Eqref{eq:stable_parameterization}, the coefficient bounds in
\Eqref{eq:coefficient_bounds} imply
\begin{align}
    g(B)
    &=
    \alpha-\left(s+2|\gamma|B\right)B+\gamma B^2
    \nonumber \\
    &\leq
    \left(sB+|\gamma|B^2\right)
    -sB-2|\gamma|B^2+\gamma B^2
    \nonumber \\
    &=
    \left(\gamma-|\gamma|\right)B^2
    \leq 0.
    \label{eq:upper_boundary}
\end{align}
At the lower boundary \(x=-B\),
\begin{align}
    g(-B)
    &=
    \alpha+\left(s+2|\gamma|B\right)B+\gamma B^2
    \nonumber \\
    &\geq
    -\left(sB+|\gamma|B^2\right)
    +sB+2|\gamma|B^2+\gamma B^2
    \nonumber \\
    &=
    \left(|\gamma|+\gamma\right)B^2
    \geq 0.
    \label{eq:lower_boundary}
\end{align}
Since \(\varepsilon>0\), the full vector field
\(\dot{x}=\varepsilon g(x)\) points inward or is tangent at both
boundaries. Hence, \([-B,B]\) is forward-invariant: if
\(x_0\in[-B,B]\), then \(x_t\in[-B,B]\) for all subsequent times. 
The trajectory therefore remains finite, and the corresponding
projective representation does not encounter a pole along the
trajectory.

\paragraph{Contractive state transitions.}
For every \(x\in[-B,B]\), the state Jacobian satisfies
\begin{align}
    \frac{\partial \dot{x}}{\partial x}
    &=
    \varepsilon\left(\beta+2\gamma x\right)
    \nonumber \\
    &=
    \varepsilon\left(
        -s-2|\gamma|B+2\gamma x
    \right)
    \nonumber \\
    &\leq
    -\varepsilon s
    \leq
    -\varepsilon s_{\min}.
    \label{eq:local_contraction}
\end{align}
Thus, the continuous-time dynamics are strictly contractive with
respect to the state throughout the invariant interval. Under the zero-order hold used in \Eqref{eq:step_matrix}, $\varepsilon_t$ and $s_t$ are constant within each step, and the
exact one-step flow therefore satisfies
\begin{equation}
    0
    <
    \frac{\partial x_t}{\partial x_{t-1}}
     \leq
    \exp(-\varepsilon_t s_t\Delta t)
    \leq
    \exp(-\varepsilon_t s_{\min}\Delta t)
    <
    1.
    \label{eq:step_contraction}
\end{equation}
Hence, the recurrent state transition is contractive within \([-B,B]\).

\paragraph{Real matrix-exponential branch.}
Finally, the discriminant in \Eqref{eq:expm} is strictly positive:
\begin{align}
    \frac{\beta^2}{4}-\alpha\gamma
    &\geq
    \frac{\left(s+2|\gamma|B\right)^2}{4}
    -
    \left(sB+|\gamma|B^2\right)|\gamma|
    \nonumber \\
    &=
    \frac{s^2}{4}
    >
    0.
    \label{eq:positive_discriminant}
\end{align}
It follows that
\begin{equation}
    \omega^2
    =
    \varepsilon^2
    \left(
        \frac{\beta^2}{4}-\alpha\gamma
    \right)
    \geq
    \frac{\varepsilon^2s^2}{4}
    >
    0.
    \label{eq:omega_lower_bound}
\end{equation}
This $\omega$ is real and strictly positive, and the exact matrix exponential in \Eqref{eq:expm} therefore always takes
the real hyperbolic branch. In summary, the proposed parameterization provides a bounded invariant
state interval, contractive state transitions within that interval,
pole-free trajectories, and a real-valued closed-form discretization.

\subsection{Projective lift of Riccati dynamics}
\label{app:lift}

For this subsection, suppress the state-dimension and time indices.
Consider the scalar Riccati equation
\begin{equation}
    \dot{x}
    =
    \varepsilon
    \left(
        \alpha+\beta x+\gamma x^2
    \right).
    \label{eq:app_riccati}
\end{equation}

\paragraph{Projective lift.}
\label{prop:lift}
Let \(p(t)\) and \(q(t)\) satisfy
\begin{equation}
    \frac{d}{dt}
    \begin{pmatrix}
        p \\ q
    \end{pmatrix}
    =
    L
    \begin{pmatrix}
        p \\ q
    \end{pmatrix},
    \qquad
    L
    =
    \varepsilon
    \begin{pmatrix}
        \beta/2 & \alpha \\
        -\gamma & -\beta/2
    \end{pmatrix}.
    \label{eq:app_lift}
\end{equation}
If \(q(t)\neq 0\) on an interval \(I\), then
\(x(t)=p(t)/q(t)\) satisfies \Eqref{eq:app_riccati} on \(I\).

\paragraph{Proof.}
Expanding \Eqref{eq:app_lift} componentwise gives
\begin{equation}
    \dot{p}
    =
    \varepsilon
    \left(
        \frac{\beta}{2}p+\alpha q
    \right),
    \qquad
    \dot{q}
    =
    \varepsilon
    \left(
        -\gamma p-\frac{\beta}{2}q
    \right).
    \label{eq:app_lift_components}
\end{equation}
Since \(x=p/q\), the quotient rule gives
\begin{align}
    \dot{x}
    &=
    \frac{\dot{p}q-p\dot{q}}{q^2}
    \nonumber \\
    &=
    \frac{
        \varepsilon
        \left[
            \left(\frac{\beta}{2}p+\alpha q\right)q
            -
            p\left(-\gamma p-\frac{\beta}{2}q\right)
        \right]
    }{q^2}
    \nonumber \\
    &=
    \varepsilon
    \left(
        \alpha+\beta\frac{p}{q}
        +\gamma\frac{p^2}{q^2}
    \right)
    \nonumber \\
    &=
    \varepsilon
    \left(
        \alpha+\beta x+\gamma x^2
    \right).
\end{align}
which recovers \Eqref{eq:app_riccati}.

The projective representation is unchanged under a common nonzero
scaling of $p$ and $q$. More generally, adding a scalar multiple of
the identity to $L$ only introduces a common multiplicative factor in
the lifted coordinates and therefore leaves the ratio $x=p/q$
unchanged. We choose the traceless representative in
\Eqref{eq:app_lift}, for which
$\operatorname{tr}(L)=0$. This choice yields the convenient identity
$L^2=\omega^2 I$ used in~\Eqref{eq:expm}.

\subsection{Exact matrix exponential}
\label{app:matrix_exponential}

Under the zero-order hold, the coefficients
\((\alpha,\beta,\gamma,\varepsilon)\) are constant over one time
interval of duration \(\Delta t\). The lifted state is therefore
updated by
\[
    z_t
    =
    \exp(\Delta t L_t)z_{t-1},
    \qquad
    z_t =
    \begin{pmatrix}
        p_t \\ q_t
    \end{pmatrix}.
\]

\paragraph{Closed-form step matrix.}
\label{prop:matrix_exponential}
Let \(L\) be the generator in \Eqref{eq:app_lift}, and define the
characteristic rate
\begin{equation}
    \omega
    =
    \sqrt{
    \varepsilon^2
    \left(
        \frac{\beta^2}{4}-\alpha\gamma
    \right)}.
    \label{eq:app_omega}
\end{equation}
If \(\omega\neq 0\), the exact step matrix is
\begin{equation}
    \exp(\Delta t L)
    =
    \cosh(\omega\Delta t)I
    +
    \frac{\sinh(\omega\Delta t)}{\omega}L.
    \label{eq:app_expm}
\end{equation}

\paragraph{Proof.}
Direct multiplication gives
\begin{equation}
    L^2
    =
    \varepsilon^2
    \begin{pmatrix}
        \beta/2 & \alpha \\
        -\gamma & -\beta/2
    \end{pmatrix}^2
    =
    \varepsilon^2
    \left(
        \frac{\beta^2}{4}-\alpha\gamma
    \right)I
    =
    \omega^2 I.
    \label{eq:app_l_squared}
\end{equation}
Thus, even powers of \(L\) are scalar multiples of \(I\), and odd
powers are scalar multiples of \(L\). Substituting these identities
into the power-series definition of the matrix exponential gives
\begin{align}
    \exp(\Delta t L)
    &=
    \sum_{k=0}^{\infty}
    \frac{(\Delta t L)^k}{k!}
    \nonumber \\
    &=
    \sum_{k=0}^{\infty}
    \frac{(\omega\Delta t)^{2k}}{(2k)!}I
    +
    \frac{1}{\omega}
    \sum_{k=0}^{\infty}
    \frac{(\omega\Delta t)^{2k+1}}{(2k+1)!}L
    \nonumber \\
    &=
    \cosh(\omega\Delta t)I
    +
    \frac{\sinh(\omega\Delta t)}{\omega}L,
\end{align}
which proves \Eqref{eq:app_expm}.

The stable parameterization in Section~\ref{sec:stability} guarantees
$\beta^2/4-\alpha\gamma>0$, and since $\varepsilon\in(0,1)$,
it follows that \(\omega^2>0\). Thus the constrained RiccatiSSM always
uses the real hyperbolic form in~\Eqref{eq:app_expm}.

\subsection{Composition of Möbius updates}
\label{app:moebius_composition}

\paragraph{Möbius composition.}
\label{prop:moebius_composition}
For a matrix
\[
    M =
    \begin{pmatrix}
        a & b \\
        c & d
    \end{pmatrix},
\]
define the associated fractional-linear map
\begin{equation}
    f_M(x)
    =
    \frac{ax+b}{cx+d},
    \label{eq:app_moebius_action}
\end{equation}
whenever \(cx+d\neq 0\). For any two matrices \(M_1\) and \(M_2\),
\begin{equation}
    f_{M_2}\left(f_{M_1}(x)\right)
    =
    f_{M_2M_1}(x)
    \label{eq:app_moebius_composition}
\end{equation}
whenever the corresponding fractional-linear maps are defined.

\paragraph{Proof.}
Let
\[
    M_1 =
    \begin{pmatrix}
        a_1 & b_1 \\
        c_1 & d_1
    \end{pmatrix},
    \qquad
    M_2 =
    \begin{pmatrix}
        a_2 & b_2 \\
        c_2 & d_2
    \end{pmatrix}.
\]
Substituting \(f_{M_1}(x)\) into \(f_{M_2}\) gives
\begin{align}
    f_{M_2}\left(f_{M_1}(x)\right)
    &=
    \frac{
        a_2\left(\frac{a_1x+b_1}{c_1x+d_1}\right)+b_2
    }{
        c_2\left(\frac{a_1x+b_1}{c_1x+d_1}\right)+d_2
    }
    \nonumber \\
    &=
    \frac{
        (a_2a_1+b_2c_1)x+(a_2b_1+b_2d_1)
    }{
        (c_2a_1+d_2c_1)x+(c_2b_1+d_2d_1)
    }.
\end{align}
These coefficients are exactly the entries of
\(M_2M_1\), which proves \Eqref{eq:app_moebius_composition}.

Repeated application of \Eqref{eq:app_moebius_composition} therefore
represents the composed updates up to time $t$ by the ordered prefix
product
\[
    P_t=M_tM_{t-1}\cdots M_1
\]
as defined in \Eqref{eq:prefix_product}. Since matrix multiplication is
associative, all prefix products can be computed with a single
associative parallel scan.

\section{Relation to Liquid Dynamics}

\paragraph{Remark: Linear lift and nonlinear state dynamics.}\label{remark:linearlift_nonlineardynamics} 
Equation~\ref{eq:lift} shows that in the lifted coordinates $(p, q)$ each channel evolves as a two-dimensional linear time-varying system with input-only coefficients, followed by the projective state $x = p/q$. This structure is what allows the sequence to be evaluated with a single scan. The nonlinearity enters through the projective representation, but $x$ is the state exposed to the next layer and follows the nonlinear Riccati dynamics in \Eqref{eq:riccati_ode}. In $x$, the dynamics are bounded, contractive, exactly integrated, and have a state-dependent local contraction rate,  and hence a state-dependent effective time constant, shown in \Eqref{eq:liquid_tc}, as in liquid models. What the lift cannot provide is nonlinear feedback into the coefficients, since making $\alpha,\beta,\gamma$ or $\varepsilon$ depend on $x$ would break closure under composition.

\subsection{Second-order local correspondence}
\label{sec:taylor_analysis}

For a fixed input \(u\), write the continuous-time vector field of
an LRC neuron as
\begin{equation}
    f_i(x;u)
    =
    \sigma\bigl(\varepsilon_i^{*}(x,u)\bigr)
    \left[
        -\sigma\bigl(f_i^{*}(x,u)\bigr)x
        +
        \tau\bigl(z_i^{*}(x,u)\bigr)e_i^{leak}
    \right],
    \label{eq:lrc_vector_field}
\end{equation}
where \(f_i^{*}\), \(z_i^{*}\), and \(\varepsilon_i^{*}\) are the LRC
pre-activations, and \(\sigma\) and \(\tau\) are its nonlinearities.
With \(u\) fixed, \(f_i(\cdot;u)\) is a scalar function of the state
\(x\).

Expanding the LRC vector field around the reference state \(x=0\)
gives
\begin{equation}
    f_i(x;u)
    =
    a_i(u)
    +
    b_i(u)x
    +
    c_i(u)x^2
    +
    \mathcal{O}(|x|^3),
    \label{eq:lrc_taylor_expansion}
\end{equation}
where
\begin{equation}
    a_i(u)
    =
    f_i(0;u),
    \qquad
    b_i(u)
    =
    f_i'(0;u),
    \qquad
    c_i(u)
    =
    \frac{1}{2}f_i''(0;u).
    \label{eq:lrc_taylor_coefficients}
\end{equation}
The corresponding quadratic field
\begin{equation}
    g_i(x;u)
    =
    a_i(u)
    +
    b_i(u)x
    +
    c_i(u)x^2
    \label{eq:matched_riccati_field}
\end{equation}
is the unique polynomial of degree at most two that matches the
LRC field's value, slope, and curvature at \(x=0\).

This quadratic field is a Riccati vector field with speed factor
\(\varepsilon_i(u)=1\). More generally, for any chosen positive speed
factor \(\varepsilon_i(u)\), the equivalent Riccati coefficients are
\begin{equation}
    \alpha_i(u)
    =
    \frac{a_i(u)}{\varepsilon_i(u)},
    \qquad
    \beta_i(u)
    =
    \frac{b_i(u)}{\varepsilon_i(u)},
    \qquad
    \gamma_i(u)
    =
    \frac{c_i(u)}{\varepsilon_i(u)}.
    \label{eq:lrc_riccati_gauge}
\end{equation}
Only the products
\(\varepsilon_i\alpha_i\),
\(\varepsilon_i\beta_i\), and
\(\varepsilon_i\gamma_i\) determine the vector field. Thus, the
decomposition into a speed factor and the remaining coefficients is
a parameterization choice.

The correspondence is local and does not imply global agreement
between the models. In particular, the LRC field is globally bounded
by its saturating nonlinearities, whereas an unconstrained quadratic
field can develop a pole outside the local operating region. The
trained RiccatiSSM uses the constrained parameterization described
in Section~\ref{sec:stability} to prevent this behavior.

The next section empirically evaluates the
accuracy and range of validity of this local correspondence,
including the error introduced by the explicit-Euler discretization
used by LrcSSM.

\subsection{Matched-dynamics experiment}
\label{sec:dyn_setup}

We next compare the continuous-time dynamics of the two models while
holding their local vector fields fixed. At every time step \(t\), we
compute the Taylor coefficients
\((a_{i,t},b_{i,t},c_{i,t})\) from
\Eqref{eq:lrc_taylor_coefficients} using the current input frame
\(u_t\). We obtain the derivatives by automatic differentiation of the
implemented LRC vector field. The Riccati systems in this analysis use
the resulting matched quadratic field
\begin{equation}
    \dot{x}_i
    =
    a_{i,t}+b_{i,t}x_i+c_{i,t}x_i^2
    \label{eq:matched_riccati_ode}
\end{equation}
instead of the learned RiccatiSSM coefficients. With this design, we aim to isolate the difference between the LRC and quadratic vector fields from
differences in parameterization and training. So with these experiments, we aim to answer whether the two models express the same dynamics.

We compare four systems on identical input sequences:
\begin{itemize}
    \item \textbf{LrcSSM-RK4} integrates the full LRC vector field
    \Eqref{eq:lrc_vector_field} using fourth-order Runge--Kutta with
    \(64\) substeps per model time step. We use this high-accuracy
    numerical solution as a reference trajectory.

    \item \textbf{LrcSSM-Euler} integrates the same LRC vector field
    with one explicit-Euler step per model time step, matching the
    discretization used by the original model.

    \item \textbf{RiccatiSSM-ZOH} integrates the matched quadratic
    field \Eqref{eq:matched_riccati_ode}. The coefficients are held
    constant within each time step, and the resulting Riccati flow is
    evaluated exactly using \Eqref{eq:expm}.

    \item \textbf{RiccatiSSM-Euler} integrates the same matched
    quadratic field with one explicit-Euler step per model time step.
\end{itemize}

These four systems distinguish approximation error from integration
error. The comparison between RiccatiSSM-ZOH and LrcSSM-RK4 measures
the combined effect of replacing the LRC field with its local
quadratic approximation while exactly integrating both held-input
systems. The comparison between LrcSSM-Euler and LrcSSM-RK4 measures
the explicit-Euler error of the original LRC dynamics. Comparing
RiccatiSSM-Euler with LrcSSM-Euler evaluates the quadratic
approximation under the same Euler integrator. 


\begin{table}[t]
\centering
\caption{Relative RMSE over $T = 512$ steps, receiving white-noise input.}
\label{tab:error_square}
\resizebox{\textwidth}{!}{%
\begin{tabular}{llc}
\toprule
Factor isolated & Comparison & Relative RMSE \\
\midrule
Truncation effect (Exact integration) & RiccatiSSM-ZOH \;vs.\; LrcSSM-RK4 & $1.73 \times 10^{-5}$ \\
Truncation effect (Euler integration) & RiccatiSSM-Euler \;vs.\; LrcSSM-Euler & $1.73 \times 10^{-5}$ \\
Integrator choice LrcSSM & LrcSSM-Euler \;vs.\; LrcSSM-RK4 & $7.16 \times 10^{-3}$ \\
\bottomrule
\end{tabular}
}
\end{table}

\begin{figure}
    \centering
    \includegraphics[width=0.99\linewidth]{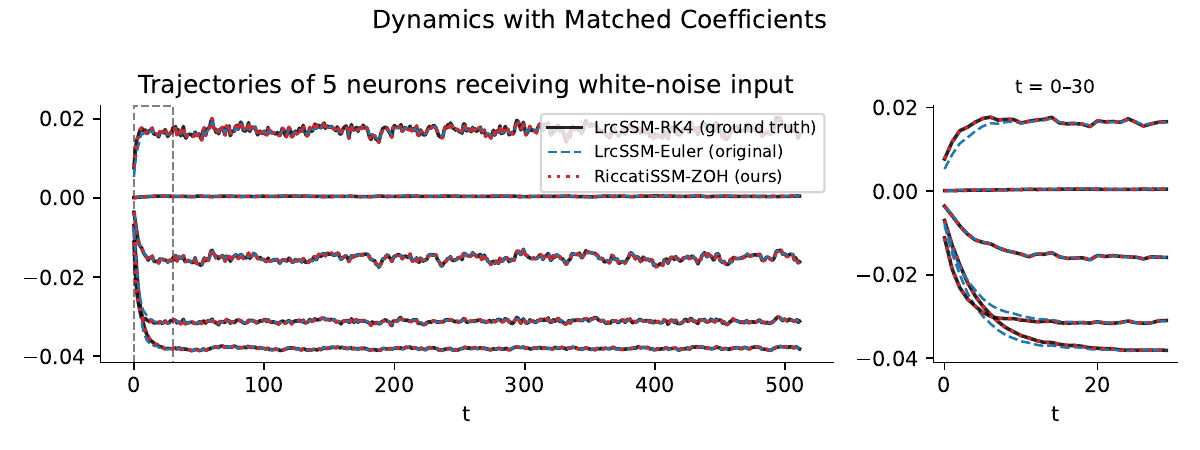}
    \caption{Hidden-state trajectories of 5 neurons under three models with matched Taylor coefficients driven by white-noise input. LrcSSM-RK4 (black, 64 RK4 substeps) serves as ground truth for the full dynamics; LrcSSM-Euler (blue dashed) is the model's single Euler step as in~\citep{farsang2025parallelization}; RiccatiSSM-ZOH (red dotted) solves the quadratic (Riccati) truncation exactly via zero-order hold. The inset shows the first 30 timesteps (dashed box). Close match between RiccatiSSM-ZOH and both LrcSSM variants indicates that the dominant discretisation error comes from the Euler integrator rather than the quadratic truncation of the vector field.
}
    \label{fig:matchedparam_dynamics}
\end{figure}

Over 512 steps, this divergence is 
$1.7\cdot 10^{-5}$ in relative RMSE, roughly two orders of magnitude smaller than the 
$7.2\cdot10^{-3}$  error that LrcSSM already incurs by discretizing its own equation with an explicit Euler step. Truncating the dynamics at second order is therefore not the limiting approximation in this model: RiccatiSSM can track the continuous LRC equation more closely than LrcSSM's own update does. 

\subsection{Reachable constraint-coefficient space}

\begin{figure}
    \centering
    \includegraphics[width=0.99\linewidth]{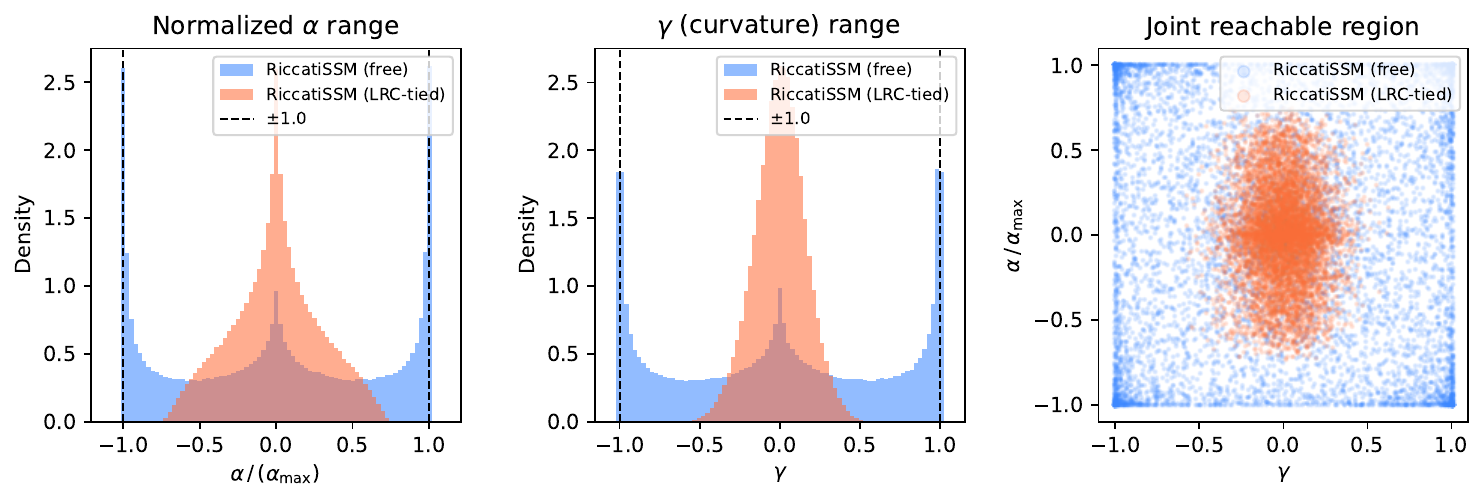}
    \caption{Reachable constraint-coefficient space under free vs. LRC-tied Riccati parameterizations, sampled over 500k random initializations. (Left) Distribution of normalized stability coefficient $\alpha/\alpha_{max}$: the free variant saturates at $\pm 1$, while the LRC-tied variant concentrates near zero due to double-squashing through the circuit nonlinearities. (Center) Distribution of curvature coefficient $\gamma$: LRC-tied outputs are similarly compressed relative to the free range. (Right) Joint ($\gamma, \alpha/\alpha_{max}$) reachable region; the LRC-tied parameterization occupies a strictly smaller subspace, reflecting the structural constraints imposed by the underlying neural circuit dynamics.}
    \label{fig:reachable_param}
\end{figure}

Figure~\ref{fig:reachable_param} visualizes the coefficient space induced by the free and LRC-tied Riccati parameterizations. In the LRC-tied variant, the Riccati coefficients are coupled through the Taylor-matching relations derived in Section~\ref{sec:taylor_analysis}, restricting the combinations of coefficients that can be realized. The default RiccatiSSM instead learns independent projections to the coefficients before applying the stability constraints, allowing it to explore a broader region of the admissible Riccati parameter space. This illustrates the additional flexibility of the free parameterization underlying the performance comparison in Section~\ref{sec:ablations}.

\section{Additional Experimental Details}
\label{app:experiments}
Our codebase builds on the implementations of~\cite{rusch2024oscillatory} and~\cite{farsang2025parallelization}. For the Weather dataset, we additionally build on the code from~\cite{discretax2025}. All experiments were run on NVIDIA A40 and A100 GPUs.

We performed a grid search over learning rates $\{10^{-5},10^{-4},10^{-3}\}$, hidden dimensions $\{16,64,128\}$, state-space dimensions $\{16,64,256\}$, and numbers of blocks $\{2,4,6\}$, following the search space of~\cite{walkerlog,rusch2024oscillatory}. For each dataset, we selected the configuration with the highest mean validation accuracy across five data splits generated using the same random seeds, ensuring direct comparability with the reported baselines. For the Weather dataset, we instead performed a random hyperparameter search over the same ranges as in~\citep{rusch2024oscillatory}. 

The best-performing hyperparameter configuration selected for each dataset is reported in Table~\ref{tab:hyperparams}.


\begin{table}[h]
    \centering
    \caption{Optimized hyperparameters used for the classification, PPG-DaLiA and Weather datasets.}
    \begin{tabular}{lccccc}
    \toprule
    & lr & hidden dim. & state-space dim. & number of layers\\
    \midrule
    Heart &  $10^{-3}$ & 64 & 64 & 4\\ 
    SCP1  & $10^{-4}$ & 64 & 16 & 6\\
    SCP2  & $10^{-3}$ & 16 & 16 & 2\\ 
    Ethanol & $10^{-3}$ & 64 & 16 & 6\\ 
    Motor & $10^{-3}$ & 16 & 256 & 4\\ 
    Worms & $10^{-3}$ & 16 & 16 & 4\\ 
    PPG-DaLiA & $10^{-3}$ & 64 & 64 & 6\\ 
    Weather & $6.8\cdot10^{-4}$ & 32 & 64 & 4\\ 
    \bottomrule
    \end{tabular}
    \label{tab:hyperparams}
\end{table}

\subsection{Runtime}
\label{app:runtime}

Table~\ref{tab:runtime} reports the runtime of the best-performing configuration selected for each dataset. These configurations may differ in architecture and model size, as they correspond to the reported benchmark results. RiccatiSSM achieves low runtimes across these configurations. 

RiccatiSSM is slower than LrcSSM on SCP1 and Ethanol; however, these comparisons are not architecture-matched. On both datasets, the selected RiccatiSSM configuration uses six layers, whereas the selected LrcSSM configuration uses only two. For a direct comparison with LrcSSM under an identical architecture, we refer to Section~\ref{sec:runtime} and Figure~\ref{fig:timing_comparison}.

\begin{table}[h]
    \centering
    \caption{Runtime in seconds for the models for 1000 training steps, using the best configuration for each. Values for the models are taken from~\cite{rusch2024oscillatory} and~\cite{farsang2025parallelization}.}
    \resizebox{\columnwidth}{!}{%
    \begin{tabular}{lccccccccccc}
    \toprule
         & NRDE & NCDE & Log-NCDE & LRU & S5 & Mamba & S6 & LinOSS-IMEX & LinOSS-IM & LrcSSM &RiccatiSSM\\
         \midrule
        Heart &  9539 &  1177 &  826 &  8 &  11 &  34 &  4 &  4 &  7 & 23 & 21 \\
        SCP1 & 1014 &  973 &  635 &  9 &  17 &  7 &  3 &  42 &  38 & 12 & 33 \\
        SCP2 &  1404 &   1251  &  583 &   9 &   9  &  32 &   7  &  55 &   22 & 15 & 12\\
        Ethanol & 2256&  2217&  2056&  16&  9&  255&  4 & 48&  8 & 15 & 26\\ 
        Motor  & 7616&   3778&   730 &  51 &  16&   35&   34&   128&   11 & 31 & 27\\
       Worms &  5386  & 24595 &  1956 &  94 &  31  & 122  & 68  & 37 &  90 & 33 & 23\\
       \bottomrule
    \end{tabular}
    }
    \label{tab:runtime}
\end{table}

\end{document}